\documentclass[runningheads]{llncs}

\usepackage{eccv}

\usepackage{eccvabbrv}

\usepackage{graphicx}
\usepackage{booktabs}
\usepackage{adjustbox}

\usepackage[accsupp]{axessibility}  

\usepackage{hyperref}

\usepackage{orcidlink}

\begin{document}

\title{Leveraging High-Resolution Features for Improved Deep Hashing-based Image Retrieval} 

\titlerunning{HHNet: HrNet for Deep Hashing}

\author{Aymene Berriche\inst{1,2}\orcidlink{0009-0007-9366-3354} \and
Mehdi Adjal Zakaria\inst{1,2}\orcidlink{0009-0008-2241-7866} \and
Riyadh Baghdadi\inst{1}\orcidlink{0000-0002-9350-3998}}

\authorrunning{A.~Berriche et al.}

\institute{New York University Abu Dhabi, United Arab Emirates  \and
École nationale supérieure d’informatique , Algeria
}

\maketitle

\begin{abstract}
Deep hashing techniques have emerged as the predominant approach for efficient image retrieval. Traditionally, these methods utilize pre-trained convolutional neural networks (CNNs) such as AlexNet and VGG-16 as feature extractors. However, the increasing complexity of datasets poses challenges for these backbone architectures in capturing meaningful features essential for effective image retrieval. In this study, we explore the efficacy of employing high-resolution features learned through state-of-the-art techniques for image retrieval tasks. Specifically, we propose a novel methodology that utilizes High-Resolution Networks (HRNets) as the backbone for the deep hashing task, termed High-Resolution Hashing Network (HHNet). Our approach demonstrates superior performance compared to existing methods across all tested benchmark datasets, including CIFAR-10, NUS-WIDE, MS COCO, and ImageNet. This performance improvement is more pronounced for complex datasets, which highlights the need to learn high-resolution features for intricate image retrieval tasks. Furthermore, we conduct a comprehensive analysis of different HRNet configurations and provide insights into the optimal architecture for the deep hashing task.

  \keywords{Deep Hashing \and High-Resolution networks \and Image Retrieval}
\end{abstract}

\section{Introduction}
\label{sec:intro} 

The rise of big data has significantly increased the volume and complexity of media data, particularly images, prompting an urgent demand for efficient and scalable retrieval methods. From traditional approaches, such as indexing methods \cite{lew2006content} and Locality-Sensitive Hashing \cite{andoni2014beyond, gionis1999similarity} to more recent state-of-the-art techniques like Deep-Hashing Networks \cite{luo2023survey}, all have played crucial roles in addressing the challenge of constructing relatively fast and efficient image retrieval algorithm for large databases. Hashing methods, in particular, have been instrumental in transforming high-dimensional media data into compact binary codes and generating similar codes for analogous data items. H owever, the emergence of large-scale and increasingly complex media data has highlighted the limitations of these conventional methods in ensuring both high retrieval quality and computational efficiency.

In response to these limitations, there has been a burgeoning interest in data-aware solutions, \ie methods that make use of the underlying structure of the given data, particularly those harnessing deep learning techniques to improve the quality of hashing methods such as \cite{xia2014supervised, liu2016deep, zhu2016deep, cao2017hashnet}. Data-aware approaches recognize the inherent structure and characteristics of the data being hashed, enabling more tailored and effective retrieval strategies. Deep learning has emerged as a potent tool in this endeavor, as it can autonomously learn intricate patterns and representations directly from raw data. By leveraging deep neural networks, we can extract high-level features that capture semantic information and relationships between images, thereby enhancing retrieval quality. Additionally, deep learning frameworks offer flexibility in handling high-dimensional and complex data, making them well-suited for the challenges posed by large-scale image databases.

Recent advancements in the field have led to the development of data-aware methodologies, significantly improving outcomes for image retrieval tasks. These methods can be divided into mainly supervised and unsupervised learning to hash methods. In unsupervised learning to hash \cite{luo2023survey,ryali2020bio}, the focus is solely on feature information or the image itself, while supervised hashing methods leverage additional label or class information to capture semantic relationships between data points \cite{cao2018deep,cao2017hashnet,xu2022hyp2, liu2016deep}. Regardless of the type, with the significant strides made by deep learning in image tasks and the use of feature extractors in pretrained CNN models such as Alexnet \cite{krizhevsky2012imagenet}, VGG-11/16 \cite{simonyan2014very} \etc. Deep hashing methods have become state-of-the-art solutions, providing more accurate and efficient retrieval in the field of image retrieval. Moreover, recent advancements have introduced additional enhancements, notably the integration of an orthogonal transformation step \cite{schwengber2023deep}. This step aims to mitigate the loss of information inherent in the transition from continuous representations learned by deep learning models to binary representations. Further exploration of these advancements will be detailed in the next section.

While traditional CNNs have demonstrated effectiveness as feature extractors not only in image retrieval but also in various other computer vision tasks such as object or pose detection \cite{nguyen2022combined, tang2022hrtransnet, zhao2019object}, a novel model architecture, known as High-resolution Networks (HrNets)  \cite{sun2019deep} has emerged as a promising alternative. An architecture that has been designed to maintain a high-resolution representation of the input through the whole process of the encoding, and has shown considerable efficacy across a spectrum of computer vision tasks \cite{wang2020deep}. This prompts an intriguing question: is learning high-resolution representations equally vital for similarity search in image retrieval? This query warrants exploration and forms a significant avenue for our research. In summary, our primary contributions encompass the following:
\begin{itemize}
    \item Firstly, we introduce a novel methodology employing HrNets to acquire high-resolution features in conjunction with the deep hashing losses, yielding state-of-the-art results in the task of image retrieval on all the tested benchmarks: NUS WIDE, MS COCO, CIFAR 10 and ImageNet datasets.
    \item Secondly, we systematically investigate various sizes of HrNets with established deep hashing techniques, aiming to discern the significance of high-resolution features in facilitating similarity search on images.
\end{itemize}

\section{Related Work}

Research in image retrieval tasks can be broadly categorized into two main classes: Traditional methods, such as Locality-Sensitive Hashing (LSH) \cite{andoni2014beyond, gionis1999similarity}, employ compact binary hash codes to represent images, enabling efficient search operations based on similarity within the hash space rather than the original image space. However, these approaches, while effective in certain scenarios, often suffer from suboptimal performance due to their agnostic nature towards the underlying data characteristics. Nevertheless, advancements in the field have demonstrated the potential for improved performance by learning hash functions directly from the data at hand. Numerous learning to hash techniques \cite{wang2017survey} have emerged in response to this need, such as Spectral Hashing \cite{weiss2008spectral} and Semi-Supervised Hashing (SSH) \cite{wang2010semi}, both build on top of principal component analysis (PCA) to construct data-aware embeddings, subsequently binarized via sign function to generate hash codes.

The second class of methods comprises Deep Hashing or Deep Learning-based hashing techniques, which are data-aware approaches. These methods enable the capture of deeper features, especially crucial for complex datasets. Subsequently, these representations are binarized to generate hash codes for nearest neighbor search in image retrieval tasks. These techniques can be further categorized into two subtypes: deep supervised hashing and deep unsupervised hashing.

The design of supervised deep hashing methods generally involves two main steps. First, the architecture is designed, which typically depends on the complexity of the task. Simple datasets like MNIST \cite{lecun1998gradient} and CIFAR-10 \cite{krizhevsky2009learning} often require a shallow architecture similar to AlexNet, whereas more complex datasets like ImageNet \cite{deng2009imagenet} and MS-COCO \cite{lin2014microsoft} necessitate a deeper network. The second phase involves designing a loss objective to train the model, where methods differ significantly. While the overarching objective is to minimize the similarity difference between the original image space and the Hamming space, variations arise in how to obtain the similarities of the learned hash codes. This categorizes supervised deep hashing methods into four main categories: pairwise methods, ranking-based methods, pointwise methods, and quantization \cite{luo2023survey}.

Typically, these methods incorporate pre-trained convolutional neural network (CNN) architectures, such as AlexNet, VGG-11, and VGG-16, followed by a sequence of fully connected layers for fine-tuning. In addition, quantization techniques have been proposed. Most methods add a quantization term to the loss during fine-tuning to account for the lossy compression done in the binarization step \cite{cao2018deep, cao2017hashnet, xu2022hyp2, zhu2016deep}. For instance, Deep Supervised Hashing (DSH) \cite{liu2016deep} employs a loss function comprising a similarity term alongside a penalization term based on $L_1$ loss. Similarly, HashNet \cite{cao2017hashnet} provides an architecture with convergence guarantees, addressing gradient instability and data imbalance issues by employing a hyperbolic tangent function to approximate the sign function during binarization. Recent work by Schwengber et al. \cite{schwengber2023deep} has proposed to split similarity learning into two steps: first, learning continuous embeddings through training the existing deep-hashing methods without accounting for the quantization penalty term, and then learning an orthogonal transformation that, when applied to the obtained embeddings, minimizes the distance between the embeddings and their binary representation.

Other notable methods include Convolutional Neural Network Hashing (CNNH) \cite{xia2014supervised}, one of the first pioneering approaches, which first derives a binary encoding approximating the data point similarities and then trains a CNN to map original data points into this binary encoding. Additionally, Deep Cauchy Hashing (DCH) \cite{cao2018deep} and its variants \cite{cao2018deepv2, kang2019maximum, li2015feature} differ primarily in their choice of loss terms for similarity and penalization. Methods employing pairwise similarity losses, such as Pairwise Correlation Discrete Hashing (PCDH) \cite{chen2020deep} and Deep Supervised Discrete Hashing (DSDH) \cite{li2017deep}, have been proposed to focus on how well hash codes can reconstruct available labels using a classifier.

Proxy-based methods at the other end, exemplified by OrthoHash, aim to maximize the cosine similarity between data points and predefined target hash codes associated with each class. Recently, HyP$^2$ \cite{xu2022hyp2} combined a proxy-based loss with a pairwise similarity term, leveraging the strengths of both approaches to achieve state-of-the-art performance.

Unsupervised hashing methods have garnered widespread attention due to their effective utilization of unlabeled data, enabling practical applications in real-world scenarios. Since deep unsupervised methods lack access to label information, semantic information is obtained in deep feature space with pre-trained networks. Methods like DistillHash \cite{yang2019distillhash}, SSDH \cite{yang2018semantic}, and HashSIM \cite{luo2022improve} have been developed in this context. However, in our work, we focus on Deep Supervised hashing methods, pairing them with HrNets as backbone to explore the importance of high-resolution features for the task at hand.

\section{Method}
\subsection{Problem Definition and Objective Functions}
The formal definition of the deep hashing problem as given in \cite{schwengber2023deep}: Given d-dimensional data or the set of images $\{x_i\}_{i=1}^{n}$, where $x_i \in \mathbb{R}^d$ and $n$ is the number of image samples, and assuming that similarity between pairs of images $S = (s_{ij})_{i,j \in [n]}$, is defined as 1 for similar images (e.g., those belonging to the same class or category) and 0 for dissimilar ones. The objective is to learn a hash function $F : \mathbb{R}^d \rightarrow \{-1, 1\}^k$, with associated hash codes $b_i = F(x_i)$, such that the Hamming distance $d_H(b_i, b_j)$ between $b_i$ and $b_j$ is minimized for similar pairs (i.e., $s_{ij} = 1$) and maximized for dissimilar ones ($s_{ij} = 0$), where :

\begin{equation}
d_H(b_i, b_j) = \frac{k - b_i^{T} b_j}{2}
\end{equation}

However it is common practice to learn a continuous embedding $f_{\theta} : \mathbb{R}^d \rightarrow \mathbb{R}^k$, with $\theta$ denoting the model parameters to be learned, and then binarize it, for example using the sign function, to get the hashing function $F = \text{sign} \circ f_{\theta}$.

To learn the optimal model parameters for hashing $\theta^{*}$, we need to optimize an objective or loss function that represents the similarity between images in terms of $d_{ij}(\theta) = d_H(F_\theta(x_i), F_\theta(x_j))$ the Hamming distance between the hashes. However, due to the discrete nature of hash functions, such objective functions are not differentiable in $\theta$. Therefore, we rely on the continuous embeddings instead of the hashes, meaning that we optimize in terms of $\tilde{d}_{ij}(\theta) = d_H(f_\theta(x_i), f_\theta(x_j))$ to get the following loss function:

\begin{equation}
    L_S(\theta) = \sum_{i,j} s_{ij} l_S \left( \tilde{d}_{ij}(\theta) \right) + (1 - s_{ij}) l_D \left( \tilde{d}_{ij}(\theta) \right)
\end{equation}

where $l_S$ and $l_D$ are losses for similar and dissimilar pairs of images in the set.

Addressing the differentiability problem has been pivotal, with numerous approaches proposed to refine the loss functions $l_S$ and $l_D$ \cite{cao2017hashnet, cao2018deep, Zhu2016, Xu2022}. However, this advancement overlooks the hashing and binarization steps during embedding learning, potentially leading to information loss upon binarization. To tackle this, various solutions have emerged, such as incorporating a quantization term into the loss function to get :
\begin{equation}
    L(\theta) = L_S(\theta) + \lambda \cdot L_Q(\theta)
\end{equation}

as suggested in prior works \cite{liu2016deep, li2016feature, tu2021}, which indeed fixes the problem but could deteriorate or hinder the neural networks' ability to learn similarity. Alternatively, a two-step learn-to-hash strategy has gained traction. Initially, continuous embeddings are learned, followed by a transformation to align them closely with their signs, ultimately enabling binarization using the sign function \cite{schwengber2023deep}, which has yielded state-of-the-art results.

\subsection{High-Resolution Hashing Networks (HHNets)}
As explained in the related works section, in typical deep hashing methodologies, convolutional neural network (CNN) architectures like AlexNet, VGG-11, and VGG-16 are commonly employed, followed by subsequent fine-tuning through fully connected layers. Our proposed approach diverges from this convention by leveraging High Resolution Networks (HRNets) first introduced in \cite{SunXLW19}, a specialized family of convolutional neural network architectures specifically tailored for tasks such as image classification, object detection, and semantic segmentation. These tasks necessitate the preservation of intricate spatial details across varying scales, a challenge that traditional CNN architectures often struggle to address adequately.

\begin{figure}[htbp]
  \centering
  \includegraphics[width=\linewidth]{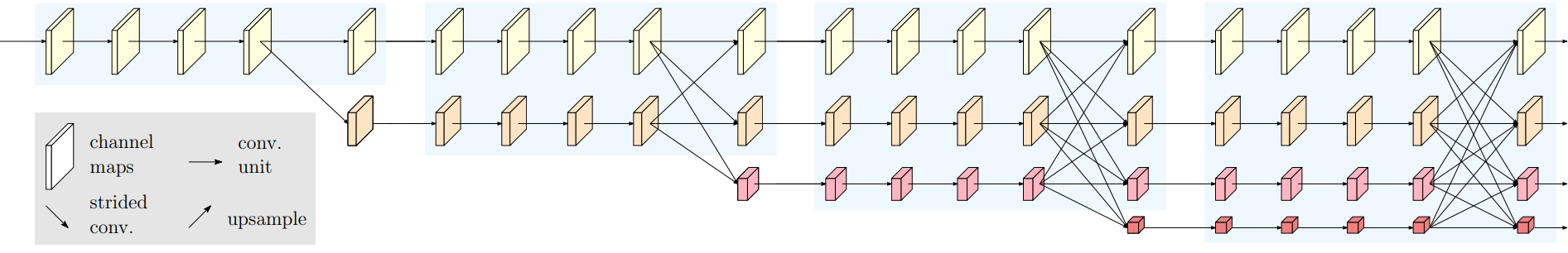}
  \caption{Illustration depicting the hierarchical structure of a high-resolution network architecture. The network consists of four stages, with the first stage employing high-resolution convolutions, while subsequent stages repeat multi-resolution blocks to handle varying levels of resolution information. This hierarchical approach ensures the preservation and effective utilization of spatial details throughout the network's processing pipeline \cite{SunXLW19, sun2019highresolution}.}
  \label{fig:hrnetbottom}
\end{figure}

The fundamental principle underlying HRNets \cite{SunXLW19} revolves around maintaining high-resolution representations throughout the architecture of the network (refer to figure \Cref{fig:hrnetbottom}). This strategy ensures the safeguarding of fine-grained spatial information that might otherwise be compromised during down-sampling operations, which are prevalent in conventional CNN designs. By prioritizing the preservation of both local and global context, HRNets demonstrate enhanced performance, particularly in scenarios involving high-resolution imagery.

\begin{figure}[htbp]
  \centering
  \includegraphics[width=0.5\linewidth]{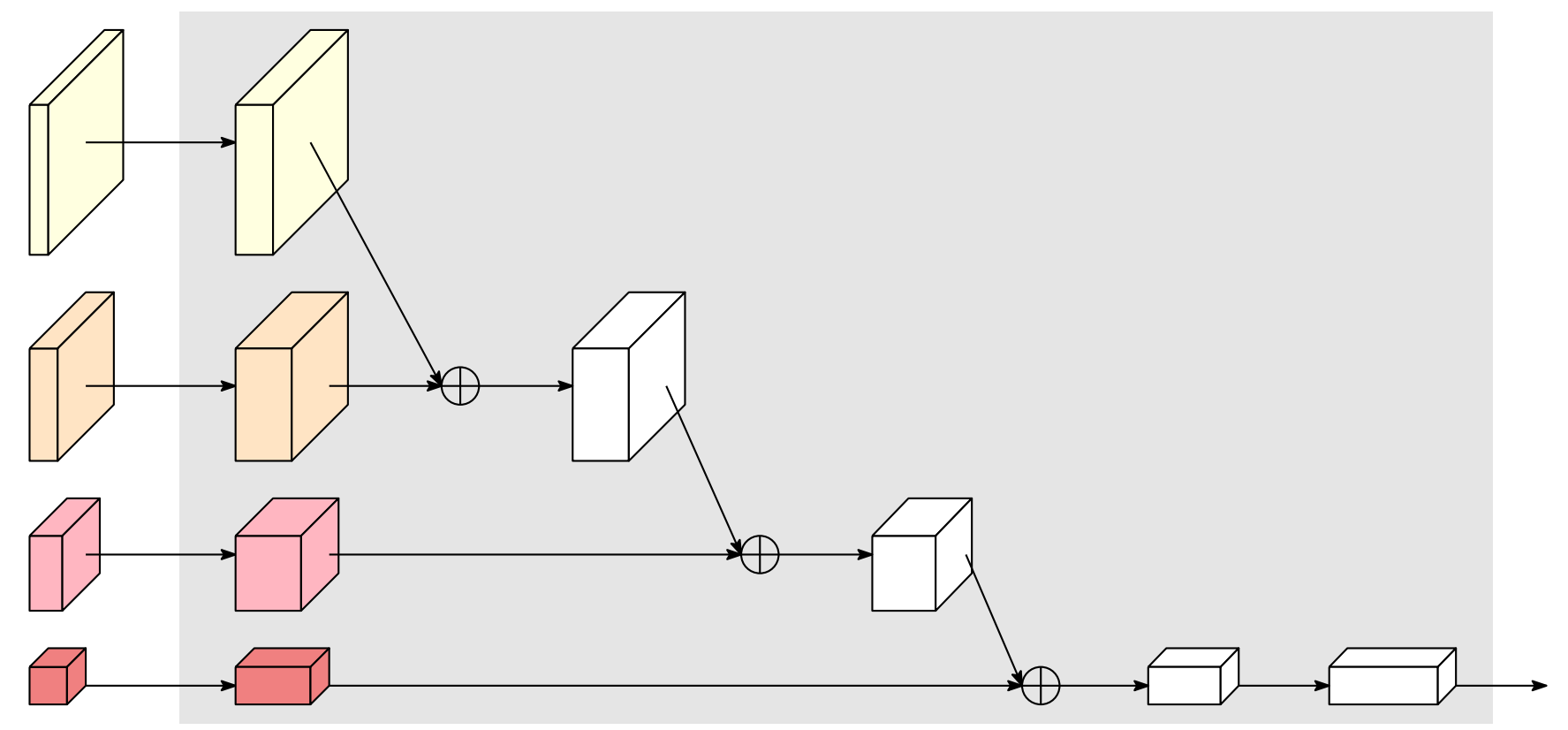}
  \caption{llustration of the augmented HRNet architecture added components \cite{sun2019highresolution}. The four-resolution feature maps are first fed into a bottleneck, increasing the number of output channels to 128, 256, 512, and 1024, respectively. Subsequently, high-resolution representations undergo downsampling through a 2-strided 3x3 convolution, outputting 256 channels, which are then added to the representations of the second-high-resolution representations. This process iterates twice to achieve 1024 channels over the smallest resolution. Finally, a transformation from 1024 to 2048 channels is accomplished through a 1x1 convolution, followed by a global average pooling operation.}
  \label{fig:AugHRNET}
\end{figure}

In a subsequent work \cite{sun2019highresolution}, which we will refer to as Augmented HRNets, a modification was introduced compared to the original HRNet approach. Unlike the original method, which only outputs the representation (feature maps) from the high-resolution convolutions \cite{SunXLW19}, Augmented HRNets make a simple yet effective adjustment (refer to \Cref{fig:AugHRNET}). They exploit other subsets of channels outputted from low-resolution convolutions. This modification enhances the capacity of the multiresolution convolution, leading to a more comprehensive exploration of the network's capabilities.

\begin{figure}[htbp]
  \centering
  \includegraphics[width=0.5\linewidth]{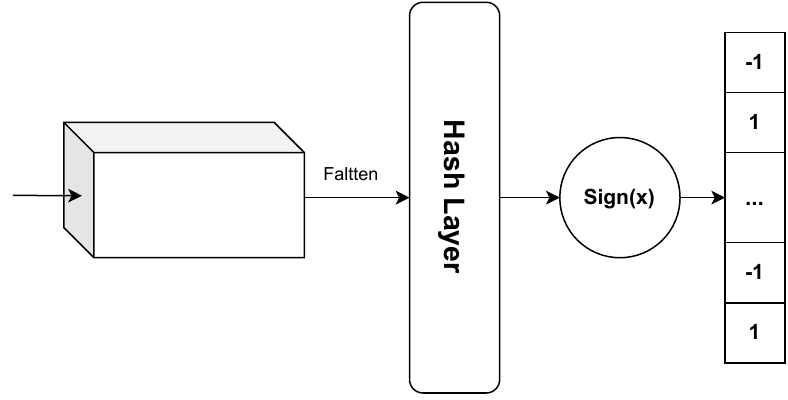}
  \caption{Figure illustrating the additional hashing layers integrated into the augmented HRNet architecture to generate image hashes from high-resolution features.}
  \label{fig:HHNET}
\end{figure}

In our work, we integrate Augmented HRNets \cite{sun2019highresolution} to be finetuned with combinations of established deep supervised hashing methods. Through this integration, we empirically demonstrate the significance of high-resolution features in the context of image retrieval tasks. Specifically, we use an augmented version of HRNets with a classification head \Cref{fig:AugHRNET} but instead of feeding the resulting 2048-dimensional representation to a classifier, it is directed to a hash layer to learn hashes \Cref{fig:HHNET}. 

Through meticulous experimentation and evaluation, we demonstrate the profound impact of HRNet architectures on the efficacy and robustness of deep supervised hashing techniques. Notably, significant enhancements in performance across all benchmarks are observed through the utilization of HRNet architectures, culminating in unprecedented results on the ImageNet dataset and state-of-the-art achievements across various benchmarks. This notable progress underscores the unmatched capabilities of HRNet in preserving fine spatial details and capturing intricate features, even within complex datasets, thus establishing a new benchmark for excellence in image retrieval.

\section{Experiments}

In the following section, we present our experimental findings. Intensive tests were conducted over various popular image retrieval benchmarks, employing a range of state-of-the-art deep hashing losses to evaluate the impact of high-resolution features. Our comparative analysis includes the two most commonly utilized backbones in the literature, namely, AlexNet and VGG-16. The objective is to assess the influence of learning high-resolution features on the quality of the acquired hashes.

Our findings can be summarized as follows:

\begin{itemize}
    \item The conducted experiments demonstrate a notable advantage in terms of precision, as indicated by the employed metric, mean average precision (\texttt{mAP}), across all benchmarks, establishing our method as a new state-of-the-art approach.
    \item The improvement in hashing quality can be further observed in complex datasets like ImageNet, where the performance increase is nearly doubled.

    \item Experiments were conducted using various HRNet model sizes, revealing that larger backbones generally yield superior hashing performance. However, the difference in performance between HRNet-W48 and HRNet-W64 is negligible.
    \item Small models like HRNet-W18 can outperform other widely utilized backbones when dealing with complex datasets such as ImageNet. This finding further emphasizes the importance of exploring the effectiveness of small high-resolution models for the deep hashing task.

\end{itemize}

In the following section, we provide details regarding our experimental setup, including the datasets utilized and other pertinent information.

\subsection{Datasets}

In accordance with the research conducted by \cite{schwengber2023deep}, and to maintain consistency and accuracy in our comparison, we utilized four image retrieval datasets:

\begin{itemize}

\item \textbf{CIFAR 10} \cite{krizhevsky2009learning}: This dataset comprises 60,000 32x32 color images distributed across 10 classes, with each class containing 6,000 images. It is widely employed for benchmarking various computer vision tasks. Consistent with prior literature \cite{cao2018deep, zhu2016deep, schwengber2023deep}, the data is partitioned to include 500 images per class in the training set, while the test and validation sets each contain 100 images per class. The remaining data is allocated as a database for querying.\\

\item \textbf{NUS WIDE} \cite{chua2009nus}: This dataset comprises 269,648 images with 5,018 tags sourced from Flickr. The images are manually annotated with 81 concepts, encompassing objects and scenes, making it a valuable resource for various computer vision tasks. Following the methodology outlined in \cite{xu2022hyp2, schwengber2023deep}, the dataset is filtered to include only images containing at least one concept from the 21 most prevalent concepts. This results in a subset of 148,332 images, from which 10,500 images are randomly sampled for the training set and 2,100 for the test and validation set. The remaining images are used as a database.\\

\item \textbf{MS COCO} \cite{lin2014microsoft}: The Microsoft Common Objects in Context (MS COCO) dataset is extensively utilized for object detection, segmentation, and captioning tasks. We use the 2017 release version, comprising 123,287 images. Following prior works \cite{cao2018deep, schwengber2023deep}, 10,000 images are sampled for the training set and 5,000 for each test and validation set, with the remaining images allocated for use as a database.\\

\item \textbf{ImageNet} \cite{deng2009imagenet}: ImageNet is a vast image database organized according to the WordNet hierarchy, with each node represented by numerous images. Widely employed in computer vision and deep learning research, it contains over 1,200,000 images. In line with prior research \cite{cao2017hashnet,schwengber2023deep}, we retain only 100 classes from the initial 1,000 possible categories. The training set comprises 13,000 images, while the test set is divided into two halves, each containing 2,500 images for querying and validation, respectively, with the remaining images designated for use as a database.

\end{itemize}

\subsection{Evaluation Metric}
We have utilized the mean average precision (\texttt{mAP}), which is widely employed as a metric in retrieval-based systems \cite{singh2022learning}. It is defined as follows:

\begin{equation}
    \texttt{mAP@k}=\frac{1}{|Q|} \sum_{q \in Q} A P_k(q)
\end{equation}
where $Q$ is the given set of queries. $AP_k$ is the average precision of the first $k \leq n$ entries \cite{schwengber2023deep}.

As done in the literature \cite{cao2017hashnet, schwengber2023deep}, we compute the \texttt{mAP@k}  with $k$ = 1000 for CIFAR 10, NUS WIDE and MS COCO and with $k$ = 5000 for ImageNet.

\subsection{Experiment Setup}
We initially investigate the effect of high-resolution features by comparing the results obtained from our adapted architecture, namely High-resolution Hashing Network (HHNet), with AlexNet and VGG-16, the two most widely used feature detectors in the deep hashing literature. As reported in \Cref{tab:full_benchmark} and \Cref{tab:hrnet_size_benchmark}, we evaluate the results using several state-of-the-art loss functions: CEL \cite{bromley1993signature}, DCH \cite{cao2018deep}, DHN \cite{zhu2016deep}, DPSH \cite{li2015feature}, Hashnet \cite{cao2017hashnet}, WGLHH \cite{tu2021weighted}, and HyP$^2$ \cite{xu2022hyp2}.

More formally, for each dataset and for each loss function (deep hashing method), and for hashing bit size $k \in \{16, 32, 48, 64\}$, we conduct training for the deep hashing method using each backbone architecture. Specifically, for AlexNet and VGG-16, we rely on the implementation provided by \cite{schwengber2023deep}, which is based on the original implementations of these methods. For HHNet, we utilize the HRNet-W64-C, an augmented version of HrNet for classification purposes that we use as a backbone. The pre-trained model can be obtained from the official implementation \href{https://github.com/HRNet/HRNet-Image-Classification}{repository}.

Furthermore, in \Cref{tab:hrnet_size_benchmark}, we examine the impact of varying the size of the HrNet backbone on the quality of hashing. We vary the model sizes and report the results accordingly. For each dataset and each of the selected model sizes, we train each deep hashing method using the designated backbone, with $k=32$. \Cref{tab:Hrnet_model_desc} shows some specifications for each backbone used in the second experiment.

\begin{table*}[]
\centering
\caption{Number of Parameters and GFLOPs as described in \cite{SunXLW19} for each of the used backbones.}
\label{tab:Hrnet_model_desc}
\adjustbox{max width = \textwidth}{
\begin{tabular}{|l|c@{\hskip .08in}|c@{\hskip .08in}|c@{\hskip .08in}|c|c@{\hskip .08in}|c@{\hskip .08in}|c@{\hskip .08in}|c|} \hline 
Backbone& \multicolumn{4}{|c|}{\# Params} & \multicolumn{4}{|c|}{GFLOPs} \\ \hline
HRNet-W18-C& \multicolumn{4}{|c|}{21.3M}& \multicolumn{4}{|c|}{1.49}\\[.5em] \hline 
HRNet-W32-C& \multicolumn{4}{|c|}{41.2M}& \multicolumn{4}{|c|}{3.99}\\[.5em] \hline 
HRNet-W48-C& \multicolumn{4}{|c|}{77.5M}& \multicolumn{4}{|c|}{16.1}\\[.5em] \hline 
HRNet-W64-C& \multicolumn{4}{|c|}{128.1M}& \multicolumn{4}{|c|}{26.9}\\[.5em] \hline\end{tabular}}

\end{table*}

\subsection{Hyperparameters}
We adopt the hyperparameter settings from \cite{schwengber2023deep}, which include a learning rate of $10^{-5}$ for fine-tuning pre-trained layers and $10^{-5}$ for the hashing layer. We utilize the Adam optimizer for optimization purposes, we train each model with a batch size of 128 for 20 epochs. The experiments were executed four times, and the results were recorded as the mean of these four runs.

\begin{table*}[htbp]

    \caption{\texttt{mAP@k} values recorded for each benchmark, with $k = 5000$ for all datasets except ImageNet, where $k = 1000$. For HHNet, an augmented version of HrNet serves as the backbone (HRNet-W64-C). The number in bold indicates the best result among all different approaches. It is noteworthy that HHNet consistently achieves the best performance across all hashing setups.}
      \label{tab:full_benchmark}

\adjustbox{max width = \textwidth}{
  \centering
  \begin{tabular}
  {l|c@{\hskip .08in}c@{\hskip .08in}c@{\hskip .08in}c|c@{\hskip .08in}c@{\hskip .08in}c@{\hskip .08in}c|c@{\hskip .08in}c@{\hskip .08in}c@{\hskip .08in}c|c@{\hskip .08in}c@{\hskip .08in}c@{\hskip .08in}c} 

 & \multicolumn{4}{c|}{CIFAR 10} & \multicolumn{4}{c|}{NUS WIDE} & \multicolumn{4}{c|}{MS COCO} & \multicolumn{4}{c}{ImageNet} \\ \hline
    number of bits ($k$)& 16&32&48&64 & 16& 32& 48&64  & 16& 32& 48&64   & 16& 32& 48&64   \\ 
    \hline
 CEL (AlexNet)& 79.8   & 81.0& 81.7& 81.3& 79.4   & 80.3& 80.7& 80.7& 64.4   & 66.3& 67.5& 68.4& 51.8   & 52.5& 53.7&45.6\\ 
 CEL (VGG-16)& 85.1   & 85.6& 85.8& 85.7& 82.7   & 83.1& 83.3& 82.8& 73.3   & 75.7& 76.2& 76.4& 69.9   & 72.8& 74.4&72.5\\ 
 CEL (\textbf{HHNet})& 87.5& 88.5& 89.3& 89.0& 84.4& 84.4& 85.1& 84.6& 80.7& 79.3& 81.0& 80.8& 87.3& 89.2& 89.9&87.2\\
 & & & & & & & & & & & & & & & &\\
 DCH (AlexNet)& 80.2   & 80.1& 80.0& 79.8& 78.4   & 79.1& 79.1& 79.8& 63.8   & 66.2& 67.1& 66.7& 58.2   & 58.8& 58.9&60.4\\
 DCH (VGG-16)& 84.0   & 82.9& 82.8& 82.9& 83.2   & 83.6& 83.6& 83.7& 68.6   & 70.7& 69.9& 71.5& 45.4   & 54.4& 59.1&62.3\\
 DCH (\textbf{HHNet})& 87.4& 87.6& 89.6& 87.9& 83.0& 84.6& 83.7& 82.9& 76.6& 81.0& 81.3& 80.8& \textbf{91.0}& \textbf{91.5}& \textbf{91.9}&\textbf{91.8}\\
 & & & & & & & & & & & & & & & &\\
 DHN (AlexNet)& 81.2   & 81.1& 81.1& 81.3& 80.6   & 81.3& 81.6& 81.7& 66.8   & 67.3& 69.2& 69.4& 25.1   & 32.4& 35.7&38.2
\\
 DHN (VGG-16)& 86.2   & 86.6& 86.9& 86.6&  83.2   & 83.6& 83.6& 83.7& 68.6   & 70.7& 69.9& 71.5& 45.4   & 54.4& 59.1&62.3\\
 DHN (\textbf{HHNet})& \textbf{90.4}& 89.8& 89.9& 90.2& 84.0& 84.9& 84.7& 85.1& 69.2& 75.5& 74.6& 75.2& 73.1& 82.3& 86.8&87.2\\
 & & & & & & & & & & & & & & & &\\
 DPSH (AlexNet)& 81.2   & 81.2& 81.5& 81.1& 81.0   & 81.9& 82.1& 82.1& 68.0   & 71.2& 71.6& 72.4& 36.5   & 42.2& 46.0&49.9\\
 
DPSH (VGG-16)& 86.4   & 86.8& 86.5& 86.7& 84.5   & 85.0& 85.2& 85.5& 77.4   & 79.0& 79.7& 79.1& 61.6   & 68.8& 71.6&74.0\\
 DPSH (\textbf{HHNet})& 90.0& \textbf{90.6}& \textbf{90.5}& \textbf{90.3}& 84.3& 85.3& 85.6& 85.7& 76.6& 76.0& 79.2& 81.0& 88.0& 85.6& 85.3&86.7\\
 & & & & & & & & & & & & & & & &\\
 HashNet (AlexNet)& 80.8   & 82.1& 82.3& 82.3& 79.8   & 81.5& 82.2& 82.7& 62.9   & 67.3& 68.2& 70.2& 41.2   & 54.3& 58.8&62.5\\
 HashNet (VGG-16)& 85.7   & 86.1& 86.4& 86.2& 83.2   & 84.1& 84.6& 85.0& 68.3   & 72.2& 74.3& 75.5& 65.0   & 74.7& 79.2&80.8\\
 HashNet (\textbf{HHNet})& 88.6& 88.9& 89.7& 89.2& 84.3& 84.9& 84.9& 85.0& 71.1& 74.5& 75.5& 74.2& 78.6& 82.1& 85.3&87.1\\
 & & & & & & & & & & & & & & & &\\
 WGLHH (AlexNet)& 79.6 &80.0 & 80.2 & 79.4 & 79.9 & 80.7 &80.1 & 80.5 & 66.3 & 67.0 & 67.7 & 67.2 & 55.3 & 57.1 & 57.0 & 56.8\\
 WGLHH(VGG-16)& 84.5   & 83.6& 84.2 & 83.3 & 83.8 & 83.6 & 83.4 & 83.0 & 78.1 & 75.9 & 76.9 & 76.6 & 79.6 & 79.4 & 79.0 &78.2\\
 WGLHH (\textbf{HHNet})& 89.2& 87.7& 87.2& 88.2& 84.6& 84.4& 84.0& 83.8& \textbf{81.0}& 81.4& 80.4& 79.6& 88.6& 89.6& 89.0&88.7\\
 & & & & & & & & & & & & & & & &\\
 HyP$^2$ (AlexNet)&80.5 &81.1 &81.7 &81.8 &81.9 &82.5 &83.1 &83.0 &71.9 &74.1 &74.8 &74.9 &54.1 &56.9 &57.7 &56.5\\
 HyP$^2$ (VGG-16)&85.0 &85.2 &85.6 &85.7 &85.1 &85.5 &85.9 &85.9 &79.7 &82.0 &82.6 &82.2 &75.7 &77.5 &79.0 &78.8\\ 
 HyP$^2$  (\textbf{HHNet})& 87.8& 88.7& 89.4& 89.8& \textbf{86.3}& \textbf{87.6}& \textbf{87.5}& \textbf{87.9}& 79.4& \textbf{86.7}& \textbf{84.2}& \textbf{84.8}& 88.5& 91.0& 91.2&91.2\\
  
  \end{tabular}}

\end{table*}

\begin{table*}[]
\caption{\texttt{mAP@k} with k=1000 is recorded for different backbone sizes of HRNet, as detailed in \Cref{tab:Hrnet_model_desc}. The bolded number highlights the best performance among the respective model sizes.}
\label{tab:hrnet_size_benchmark}

\adjustbox{max width = \textwidth}{
\begin{tabular}{l|c@{\hskip .08in}c@{\hskip .08in}c@{\hskip .08in}c|c@{\hskip .08in}c@{\hskip .08in}c@{\hskip .08in}c|c@{\hskip .08in}c@{\hskip .08in}c@{\hskip .08in}c|c@{\hskip .08in}c@{\hskip .08in}c@{\hskip .08in}c}
& \multicolumn{4}{c|}{CIFAR 10} & \multicolumn{4}{c|}{NUS WIDE} & \multicolumn{4}{c|}{MS COCO} & \multicolumn{4}{c}{ImageNet} \\ \hline
Model& W18& W32& W48& W64& W18& W32& W48& W64& W18& W32& W48& W64& W18& W32& W48& W64\\ \hline
CEL& 80.0& 86.5& 88.7& 88.5& 84.0& 83.5& 84.1& 84.4& 79.2& 77.5& 78.7& 79.3& 87.0& 83.7& 87.1& 89.2\\[.5em]
DCH& 79.4& 84.2& 87.6& 87.6& 82.4& 81.6& 82.7& 84.6& 77.2& 77.4& 79.6& 81.0& \textbf{87.9}& \textbf{86.2}& \textbf{89.7}& \textbf{91.5}\\[.5em]
DHN& \textbf{81.3}& \textbf{87.1}& \textbf{88.6}& 89.8& 83.7& 83.6& 84.1& 84.9& 70.7& 71.5& 72.7& 75.5& 76.7& 73.7& 80.5& 82.3\\[.5em]
DPSH& 82.1& 87.0& 89.2& \textbf{90.6}& 84.2& 84.3& 85.3& 85.3& 79.1& 77.7& 79.1& 76.0& 75.5& 76.6& 83.3& 85.6\\[.5em]
HashNet& 82.0& 86.8& 88.3& 88.9& 84.1& 84.1& 84.2& 84.9& 70.1& 72.1& 73.7& 74.5& 79.2& 79.5& 81.8& 82.1\\[.5em]
WGLHH& 79.3& 84.4& 87.4& 87.7& 83.8& 83.8& 84.4& 84.4& 79.6& 79.2& 80.2& 81.4& 84.1& 85.4& 88.0& 89.6\\[.5em]
HyP$^2$& 79.7& 84.5& 88.1& 88.7& \textbf{87.2}& \textbf{86.1}& \textbf{87.1}& \textbf{87.6}& \textbf{81.5}& \textbf{81.5}& \textbf{83.1}& \textbf{86.7}& 87.0& 84.9& 88.5& 91.0\end{tabular}}
\end{table*}

\subsection{Improvement Over Deep Hashing Methods}
Results from \Cref{tab:full_benchmark} demonstrate that our proposed adaptation of HrNet outperforms both AlexNet and VGG-16 in most experiments, with improvements of up to almost two-fold in terms of mean average precision. Especially when dealing with complex datasets such as ImageNet, high-resolution features prove to be more efficient.

Furthermore, it is noteworthy that HHNet trained using the DCH and HyP$^2$ losses generally outperforms other methods in most settings, although exceptions exist such as DPSH, and WGLHH which exhibit superior performance in some experiments.

The impact of learning at high resolutions is evident in the conducted experiments. Although the improvement can be marginal in some cases, we would like to emphasize that our proposed adaptation does not introduce additional layers to be trained, unlike many state-of-the-art methods. Instead, we leverage the output features to learn a hash layer that effectively learns hashes, thereby streamlining the process.

When exploring various sizes of HRNet as the backbone (\Cref{tab:hrnet_size_benchmark}), a trend emerges indicating that larger sizes generally yield better results. However, the difference in performance can be very small between models like HRNet-W48 and HRNet-W64.

Another important observation is that while other backbones fail to perform well for complex datasets like ImageNet, relatively small HRNet models like HRNet-W18 can still be very effective and achieve state-of-the-art performance. This further opens potential avenues for adapting lightweight high-resolution models for effective deep hashing.

In the next section, we discuss this aspect further.

\section{Limitation}

While the proposed High-Resolution Hashing Network (HRNet) exhibits superior performance compared to the widely used backbones in the literature, namely AlexNet and VGG-16, it is essential to acknowledge that the size of the network impacts the efficiency of such an approach. However, this does not negate the need to learn high-resolution features to address complex datasets in the task of learning to hash. The results from \Cref{tab:hrnet_size_benchmark} demonstrate the great potential of smaller HRNet models like HRNet-W18, highlighting their capacity to perform remarkably well on complex datasets such as ImageNet. We believe that exploring the effectiveness of smaller high-resolution networks would be a promising avenue for future research. Adapting lightweight architectures like Lite-HRNet \cite{yu2021lite} for the task of deep hashing could be a viable solution to address the problem of efficiency regarding complex datasets.

\section{Conclusion}

In this work, we have proposed a High-Resolution Hashing Network (HRNet), a novel adaptation of HRNet for the task of deep hashing. The results obtained by comparing with state-of-the-art deep hashing methods demonstrate that the proposed architecture outperforms other widely utilized backbones across all datasets and achieves approximately twice the performance gain on complex datasets like ImageNet. Our study highlights the need to employ high-resolution features for complex datasets. Lastly, we emphasize the potential of exploring lightweight versions of HRNet for an effective deep hashing strategy. We hope our research encourages the community to further explore the feature representation aspect of this important task.

\section{Acknowledgement}
We sincerely thank Mr. Laissaoui Akram Badreddine for their valuable advice and ideas that were essential to this work. This research has been partly supported by the Center for Artificial Intelligence and Robotics (CAIR) at New York University Abu Dhabi. A part of the training was carried out on the High Performance Computing resources at New York University Abu Dhabi.


%
%
\bibliographystyle{splncs04}
\bibliography{main}

\begin{thebibliography}{10}
\providecommand{\url}[1]{\texttt{#1}}
\providecommand{\urlprefix}{URL }
\providecommand{\doi}[1]{https://doi.org/#1}

\bibitem{andoni2014beyond}
Andoni, A., Indyk, P., Nguyen, H.L., Razenshteyn, I.: Beyond locality-sensitive hashing. In: Proceedings of the twenty-fifth annual ACM-SIAM symposium on Discrete algorithms. pp. 1018--1028. SIAM (2014)

\bibitem{bromley1993signature}
Bromley, J., Guyon, I., LeCun, Y., S{\"a}ckinger, E., Shah, R.: Signature verification using a" siamese" time delay neural network. Advances in neural information processing systems  \textbf{6} (1993)

\bibitem{cao2018deep}
Cao, Y., Long, M., Liu, B., Wang, J.: Deep cauchy hashing for hamming space retrieval. In: Proceedings of the IEEE Conference on Computer Vision and Pattern Recognition. pp. 1229--1237 (2018)

\bibitem{cao2017hashnet}
Cao, Z., Long, M., Wang, J., Yu, P.S.: Hashnet: Deep learning to hash by continuation. In: Proceedings of the IEEE international conference on computer vision. pp. 5608--5617 (2017)

\bibitem{cao2018deepv2}
Cao, Z., Sun, Z., Long, M., Wang, J., Yu, P.S.: Deep priority hashing. In: Proceedings of the 26th ACM international conference on Multimedia. pp. 1653--1661 (2018)

\bibitem{chen2020deep}
Chen, Y., Lu, X.: Deep discrete hashing with pairwise correlation learning. Neurocomputing  \textbf{385},  111--121 (2020)

\bibitem{chua2009nus}
Chua, T.S., Tang, J., Hong, R., Li, H., Luo, Z., Zheng, Y.: Nus-wide: a real-world web image database from national university of singapore. In: Proceedings of the ACM international conference on image and video retrieval. pp.~1--9 (2009)

\bibitem{deng2009imagenet}
Deng, J., Dong, W., Socher, R., Li, L.J., Li, K., Fei-Fei, L.: Imagenet: A large-scale hierarchical image database. In: 2009 IEEE conference on computer vision and pattern recognition. pp. 248--255. Ieee (2009)

\bibitem{gionis1999similarity}
Gionis, A., Indyk, P., Motwani, R., et~al.: Similarity search in high dimensions via hashing. In: Vldb. vol.~99, pp. 518--529 (1999)

\bibitem{kang2019maximum}
Kang, R., Cao, Y., Long, M., Wang, J., Yu, P.S.: Maximum-margin hamming hashing. In: Proceedings of the IEEE/CVF international conference on computer vision. pp. 8252--8261 (2019)

\bibitem{krizhevsky2009learning}
Krizhevsky, A., Hinton, G., et~al.: Learning multiple layers of features from tiny images  (2009)

\bibitem{krizhevsky2012imagenet}
Krizhevsky, A., Sutskever, I., Hinton, G.E.: Imagenet classification with deep convolutional neural networks. Advances in neural information processing systems  \textbf{25} (2012)

\bibitem{lecun1998gradient}
LeCun, Y., Bottou, L., Bengio, Y., Haffner, P.: Gradient-based learning applied to document recognition. Proceedings of the IEEE  \textbf{86}(11),  2278--2324 (1998)

\bibitem{lew2006content}
Lew, M.S., Sebe, N., Djeraba, C., Jain, R.: Content-based multimedia information retrieval: State of the art and challenges. ACM Transactions on Multimedia Computing, Communications, and Applications (TOMM)  \textbf{2}(1),  1--19 (2006)

\bibitem{li2017deep}
Li, Q., Sun, Z., He, R., Tan, T.: Deep supervised discrete hashing. Advances in neural information processing systems  \textbf{30} (2017)

\bibitem{li2015feature}
Li, W.J., Wang, S., Kang, W.C.: Feature learning based deep supervised hashing with pairwise labels. arXiv preprint arXiv:1511.03855  (2015)

\bibitem{li2016feature}
Li, W.J., Wang, S., Kang, W.C.: Feature learning based deep supervised hashing with pairwise labels (2016)

\bibitem{lin2014microsoft}
Lin, T.Y., Maire, M., Belongie, S., Hays, J., Perona, P., Ramanan, D., Doll{\'a}r, P., Zitnick, C.L.: Microsoft coco: Common objects in context. In: Computer Vision--ECCV 2014: 13th European Conference, Zurich, Switzerland, September 6-12, 2014, Proceedings, Part V 13. pp. 740--755. Springer (2014)

\bibitem{liu2016deep}
Liu, H., Wang, R., Shan, S., Chen, X.: Deep supervised hashing for fast image retrieval. In: Proceedings of the IEEE conference on computer vision and pattern recognition. pp. 2064--2072 (2016)

\bibitem{luo2022improve}
Luo, X., Ma, Z., Cheng, W., Deng, M.: Improve deep unsupervised hashing via structural and intrinsic similarity learning. IEEE Signal Processing Letters  \textbf{29},  602--606 (2022)

\bibitem{luo2023survey}
Luo, X., Wang, H., Wu, D., Chen, C., Deng, M., Huang, J., Hua, X.S.: A survey on deep hashing methods. ACM Transactions on Knowledge Discovery from Data  \textbf{17}(1),  1--50 (2023)

\bibitem{nguyen2022combined}
Nguyen, H.C., Nguyen, T.H., Nowak, J., Byrski, A., Siwocha, A., Le, V.H.: Combined yolov5 and hrnet for high accuracy 2d keypoint and human pose estimation. Journal of Artificial Intelligence and Soft Computing Research  \textbf{12}(4),  281--298 (2022)

\bibitem{ryali2020bio}
Ryali, C., Hopfield, J., Grinberg, L., Krotov, D.: Bio-inspired hashing for unsupervised similarity search. In: International conference on machine learning. pp. 8295--8306. PMLR (2020)

\bibitem{schwengber2023deep}
Schwengber, L.R., Resende, L., Orenstein, P., Oliveira, R.I.: Deep hashing via householder quantization. arXiv preprint arXiv:2311.04207  (2023)

\bibitem{simonyan2014very}
Simonyan, K., Zisserman, A.: Very deep convolutional networks for large-scale image recognition. arXiv preprint arXiv:1409.1556  (2014)

\bibitem{singh2022learning}
Singh, A., Gupta, S.: Learning to hash: A comprehensive survey of deep learning-based hashing methods. Knowledge and Information Systems  \textbf{64}(10),  2565--2597 (2022)

\bibitem{sun2019deep}
Sun, K., Xiao, B., Liu, D., Wang, J.: Deep high-resolution representation learning for human pose estimation. In: Proceedings of the IEEE/CVF conference on computer vision and pattern recognition. pp. 5693--5703 (2019)

\bibitem{SunXLW19}
Sun, K., Xiao, B., Liu, D., Wang, J.: Deep high-resolution representation learning for human pose estimation. In: CVPR (2019)

\bibitem{sun2019highresolution}
Sun, K., Zhao, Y., Jiang, B., Cheng, T., Xiao, B., Liu, D., Mu, Y., Wang, X., Liu, W., Wang, J.: High-resolution representations for labeling pixels and regions (2019)

\bibitem{tang2022hrtransnet}
Tang, B., Liu, Z., Tan, Y., He, Q.: Hrtransnet: Hrformer-driven two-modality salient object detection. IEEE Transactions on Circuits and Systems for Video Technology  \textbf{33}(2),  728--742 (2022)

\bibitem{tu2021}
Tu, R.C., Mao, X.L., Kong, C., Shao, Z., Li, Z.L., Wei, W., Huang, H.: Weighted gaussian loss based hamming hashing. In: Proceedings of the 29th ACM International Conference on Multimedia. p. 3409–3417. MM '21, Association for Computing Machinery, New York, NY, USA (2021). \doi{10.1145/3474085.3475498}, \url{https://doi.org/10.1145/3474085.3475498}

\bibitem{tu2021weighted}
Tu, R.C., Mao, X.L., Kong, C., Shao, Z., Li, Z.L., Wei, W., Huang, H.: Weighted gaussian loss based hamming hashing. In: Proceedings of the 29th ACM International Conference on Multimedia. pp. 3409--3417 (2021)

\bibitem{wang2020deep}
Wang, J., Sun, K., Cheng, T., Jiang, B., Deng, C., Zhao, Y., Liu, D., Mu, Y., Tan, M., Wang, X., et~al.: Deep high-resolution representation learning for visual recognition. IEEE transactions on pattern analysis and machine intelligence  \textbf{43}(10),  3349--3364 (2020)

\bibitem{wang2017survey}
Wang, J., Zhang, T., Sebe, N., Shen, H.T., et~al.: A survey on learning to hash. IEEE transactions on pattern analysis and machine intelligence  \textbf{40}(4),  769--790 (2017)

\bibitem{wang2010semi}
Wang, J., Kumar, S., Chang, S.F.: Semi-supervised hashing for scalable image retrieval. In: 2010 IEEE Computer Society Conference on Computer Vision and Pattern Recognition. pp. 3424--3431. IEEE (2010)

\bibitem{weiss2008spectral}
Weiss, Y., Torralba, A., Fergus, R.: Spectral hashing. Advances in neural information processing systems  \textbf{21} (2008)

\bibitem{xia2014supervised}
Xia, R., Pan, Y., Lai, H., Liu, C., Yan, S.: Supervised hashing for image retrieval via image representation learning. In: Proceedings of the AAAI conference on artificial intelligence. vol.~28 (2014)

\bibitem{xu2022hyp2}
Xu, C., Chai, Z., Xu, Z., Yuan, C., Fan, Y., Wang, J.: Hyp2 loss: Beyond hypersphere metric space for multi-label image retrieval. In: Proceedings of the 30th ACM International Conference on Multimedia. pp. 3173--3184 (2022)

\bibitem{Xu2022}
Xu, C., Chai, Z., Xu, Z., Yuan, C., Fan, Y., Wang, J.: Hyp2 loss: Beyond hypersphere metric space for multi-label image retrieval. In: Proceedings of the 30th ACM International Conference on Multimedia. p. 3173–3184. MM '22, Association for Computing Machinery, New York, NY, USA (2022). \doi{10.1145/3503161.3548032}, \url{https://doi.org/10.1145/3503161.3548032}

\bibitem{yang2018semantic}
Yang, E., Deng, C., Liu, T., Liu, W., Tao, D.: Semantic structure-based unsupervised deep hashing. In: Proceedings of the 27th international joint conference on artificial intelligence. pp. 1064--1070 (2018)

\bibitem{yang2019distillhash}
Yang, E., Liu, T., Deng, C., Liu, W., Tao, D.: Distillhash: Unsupervised deep hashing by distilling data pairs. In: Proceedings of the IEEE/CVF conference on computer vision and pattern recognition. pp. 2946--2955 (2019)

\bibitem{yu2021lite}
Yu, C., Xiao, B., Gao, C., Yuan, L., Zhang, L., Sang, N., Wang, J.: Lite-hrnet: A lightweight high-resolution network. In: Proceedings of the IEEE/CVF conference on computer vision and pattern recognition. pp. 10440--10450 (2021)

\bibitem{zhao2019object}
Zhao, Z.Q., Zheng, P., Xu, S.t., Wu, X.: Object detection with deep learning: A review. IEEE transactions on neural networks and learning systems  \textbf{30}(11),  3212--3232 (2019)

\bibitem{zhu2016deep}
Zhu, H., Long, M., Wang, J., Cao, Y.: Deep hashing network for efficient similarity retrieval. In: Proceedings of the AAAI conference on Artificial Intelligence. vol.~30 (2016)

\bibitem{Zhu2016}
Zhu, H., Long, M., Wang, J., Cao, Y.: Deep hashing network for efficient similarity retrieval. Proceedings of the AAAI Conference on Artificial Intelligence  \textbf{30}(1) (Mar 2016). \doi{10.1609/aaai.v30i1.10235}, \url{https://ojs.aaai.org/index.php/AAAI/article/view/10235}

\end{thebibliography}
\end{document}